\documentclass[runningheads]{llncs}
\usepackage[T1]{fontenc}
\usepackage{graphicx}
\usepackage{diagbox}
\usepackage{amsmath}
\usepackage{bbm}
\usepackage{booktabs}
\usepackage{makecell}
\usepackage{multirow}
\usepackage{url}
\usepackage{bbding}
\usepackage{xspace}
\usepackage{microtype}
\usepackage{orcidlink}
\usepackage[misc]{ifsym}

\newcommand{\modelname}{DIFT\xspace}
\newcommand{\fb}{FB15K-237\xspace}
\newcommand{\wn}{WN18RR\xspace}
\newcommand{\datasampling}{truncated sampling\xspace}
\begin{document}
\title{Finetuning Generative Large Language Models with Discrimination Instructions for\\ Knowledge Graph Completion}
\titlerunning{Finetuning Generative LLMs with Discrimination Instructions}

\author{Yang Liu\inst{1}\orcidlink{0000-0002-4663-2439} \and
Xiaobin Tian\inst{1}\orcidlink{0000-0001-6570-6424} \and
Zequn Sun\inst{1}$^{(\textrm{\Letter})}$\orcidlink{0000-0003-4177-9199} \and
Wei Hu\inst{1,2}\orcidlink{0000-0003-3635-6335}
}
\authorrunning{Y. Liu et al.}
\institute{
State Key Laboratory for Novel Software Technology,\\
Nanjing University, Nanjing 210023, China 
\and
National Institute of Healthcare Data Science,\\
Nanjing University, Nanjing 210093, China\\
\email{\{yliu20, xbtian\}.nju@gmail.com, \{sunzq, whu\}@nju.edu.cn}\\
}

\maketitle              % typeset the header of the contribution
\begin{abstract}
Traditional knowledge graph (KG) completion models learn embeddings to predict missing facts. 
Recent works attempt to complete KGs in a text-generation manner with large language models (LLMs).
However, they need to ground the output of LLMs to KG entities, which inevitably brings errors.
In this paper, we present a finetuning framework, \modelname, aiming to unleash the KG completion ability of LLMs and avoid grounding errors.
Given an incomplete fact, \modelname employs a lightweight model to obtain candidate entities and finetunes an LLM with discrimination instructions to select the correct one from the given candidates.
To improve performance while reducing instruction data, \modelname uses a truncated sampling method to select useful facts for finetuning and injects KG embeddings into the LLM.
Extensive experiments on benchmark datasets demonstrate the effectiveness of our proposed framework.

\keywords{Knowledge graph completion \and Large language model \and Instruction tuning.}
\end{abstract}

%==============================
\section{Introduction}
Knowledge graphs (KGs) store real-world facts in multi-relational structures, where nodes represent entities and edges are labeled with relations to describe facts in the form of triplets like (\texttt{head entity}, \texttt{relation}, \texttt{tail entity}).
KGs often face the incompleteness problem \cite{KGC}, which adversely affects the performance of downstream knowledge-intensive applications such as question answering \cite{COKG-QA,EmbedKGQA} and recommender systems \cite{recomender_system}. 
KG completion models are designed to resolve the incompleteness issue by inferring the missing facts based on the facts already in KGs.
Conventional KG completion models are based on KG embeddings.
Given an incomplete fact where either the head or tail entity is missing and requires prediction, \textit{embedding-based models} first compute the plausibility for candidate entities using an embedding function of entities and relations and then rank them to obtain predictions.
Entity and relation embeddings can be learned based on either graph structures \cite{TuckER,TransE,LPFGF,NBFNet} or text attributes \cite{CoLE,CapsKG,StAR,SimKGC,KG-BERT}.

In recent years, motivated by the impressive performance of generative pre-trained language models (PLMs) such as T5~\cite{T5} and BART~\cite{BART}, some models convert KG completion to a sequence-to-sequence generation task \cite{KG-S2S,KGT5,GenKGC}.
Given an incomplete fact, \textit{generation-based models} first construct a natural language query with text attributes of the given entity and relation, and then ask a generative PLM to generate an answer directly.
Finally, they ground the answer to the entities in the KG, which, however, inevitably brings errors.

More recently, some work attempts to conduct KG completion using large language models (LLMs), such as ChatGPT and LLaMA~\cite{LLaMA}.
Given an incomplete fact, KICGPT \cite{KICGPT} first constructs query prompts with demonstration facts and the top-$m$ candidate entities predicted by a pre-trained KG completion model.
Then, it engages in a multi-round online interaction with ChatGPT using these query prompts.
Finally, it rearranges these candidates according to the response of ChatGPT.
This method may not make full use of the reasoning ability of LLMs because the LLMs (e.g., ChatGPT)
may not fit the KG well. 
Besides, the multi-round interaction costs too much.
In contrast, KG-LLM \cite{KG-LLM} converts KG completion queries to natural language questions and finetunes LLMs (e.g., LLaMA-7B) to generate answers.
It then uses a heuristic method to ground the output of LLMs to KG entities: if the output text contains an entity name, then this entity is selected as the answer.
The errors in such a grounding process cause KG-LLM to lag behind the state-of-the-art KG completion models.
Besides, generation-based models obtain multiple output texts and rank them by the generation probabilities, which is time-consuming and unsuitable for LLMs.

To address the above issues and fully exploit the reasoning ability of LLMs, we propose \modelname that finetunes LLMs with discrimination instructions for KG completion.
To avoid the grounding errors in generation-based models,  
\modelname constructs discrimination instructions that require LLMs to select an entity from the given candidates as the answer.
Specifically,
it first employs a lightweight embedding-based model to provide the top-$m$ candidate entities for each incomplete fact, 
and adds the names of these entities to the prompts as candidate answers to the KG completion query.
Then, it finetunes an LLM with parameter-efficient finetuning methods like LoRA \cite{LoRA} to select one entity name from the prompt as the output.
In this way, the LLM gets enhanced by finetuning and can always generate entities in the KG instead of unconstrained generation.

However, training the LLM with parameter-efficient finetuning methods is still costly.
To further reduce the computation cost of finetuning, we design a truncated sampling method that can select useful samples from the KG for instruction construction.
Let us assume that we get an example for finetuning with the query $q=(\textit{h},\textit{r}, ?)$ and the answer entity $\textit{t}$.
We use the pre-trained embedding-based model to compute the score of the fact $(\textit{h},\textit{r}, \textit{t})$ and the rank of $\textit{t}$. 
Then, the truncated sampling method decides whether to discard the example based on the score of the fact and the rank of the answer entity.
To unleash the graph reasoning ability of the LLM on KGs, 
we inject the embedded knowledge of queries and candidate entities into the LLM to further enhance it.

In summary, our main contributions are threefold:
\begin{itemize}
    \item We propose a new KG completion framework, namely \modelname, which leverages \underline{d}iscrimination \underline{i}nstructions to \underline{f}ine\underline{t}une generative LLMs. 
    \modelname does not require grounding the output of LLMs to entities in KGs.
    
    \item We propose a truncated sampling method to select useful KG samples for instruction construction to improve finetuning efficiency.
    % We also add KG embeddings into the input layer to improve finetuning effectiveness.
    We also inject KG embeddings into LLMs to improve finetuning effectiveness.
    
    \item Experiments show that \modelname advances the state-of-the-art KG completion results, with 0.364 Hits@1 on \fb and 0.616 on \wn.
\end{itemize}

The remaining sections of this paper are structured as follows.
In Section~\ref{sect:related_work}, we delve into the existing research on knowledge graph completion.
Section~\ref{sect:method} provides a detailed exposition of our proposed framework.
We then present our experimental results and analyses in Section~\ref{sect:experiment}.
Finally, in Section~\ref{sect:conclusion}, we conclude this paper and outline potential avenues for future research.

\section{Related Work}\label{sect:related_work}
Related studies can be divided into embedding- and generation-based models. 

\subsection{Embedding-based KG Completion}
Embedding-based KG completion methods compute prediction probability with entity and relation embeddings learned from either structural or textual features.
We divide existing embedding-based models into two categories: structure-based models and PLM-based models.

\medskip\noindent\textbf{Structure-based Models.}
These models learn embeddings using structural features such as edges (i.e., triplets), paths or neighborhood subgraphs.
Therefore, they can be categorized into three groups.
The first group comprises triplet-based embedding models to preserve the local relational structures of KGs.
They interpret relations as geometric transformations \cite{TransE,RotatE} or utilize semantic matching methods for scoring triplets \cite{TuckER,Dihedron}.
The second group contains path-based models \cite{NCRL,Neural-LP}, which predominantly learn probabilistic logical rules from relation paths to facilitate reasoning and infer missing entities.
The models in the third group use various deep neural networks to encode the subgraph structures of KGs.
CompGCN \cite{CompGCN} captures the semantics of multi-relational graphs of KGs based on the graph convolutional networks (GCN) framework.
Instead, HittER \cite{HittER} uses Transformer to aggregate relational neighbor information.
Recently, NBFNet \cite{NBFNet} employs a flexible and general framework to learn the representation of entity pairs, demonstrating strong performance among structure-based models.

\medskip\noindent\textbf{PLM-based Models.}
PLM-based models employ PLMs (e.g., BERT \cite{BERT}) to encode the text attributes of entities and relations in facts, and compute prediction probabilities using the output embeddings.
KG-BERT \cite{KG-BERT} is the first PLM-based KG completion model, which verifies that PLMs are capable of capturing the factual knowledge in KGs.
It turns a fact into a natural language sentence by concatenating entity and relation names, and then predicts whether this sentence is correct or not.
Following KG-BERT, some subsequent works make improvements in different aspects.
StAR \cite{StAR} divides each fact into two asymmetric parts and encodes them separately with a Siamese-style encoder.
SimKGC \cite{SimKGC} introduces three types of negatives for efficient contrastive learning.
CoLE \cite{CoLE} promotes structure-based models and PLM-based models mutually through a co-distillation learning framework.
These works are all embedding-based models. They obtain query embeddings and entity embeddings with encoder-only PLMs like BERT.

\subsection{Generation-based KG Completion}
Different from embedding-based models that need to learn entity, relation or fact embeddings,
generation-based models convert KG completion as a text generation task.
These models first translate a KG completion query into a natural language question and then ask a generative language model (e.g., T5 \cite{T5} and BART \cite{BART}) to give an answer.
Finally, they ground answers to entities in KGs using some matching methods.
To compare with traditional KG completion models that rank entities based on their scores, generation-based models generate multiple entities with beam search or sampling and rank them by the generation probabilities.
GenKGC \cite{GenKGC} converts KG completion to sequence-to-sequence generation task to achieve fast inference speed.
KGT5 \cite{KGT5} designs a unified framework for KG completion and question answering, but discards the pre-trained weights and trains T5 from scratch.
KG-S2S \cite{KG-S2S} proposes to employ a generative language model to solve different forms of KG completion tasks including static KG completion, temporal KG completion, and few-shot KG completion \cite{KG_survey}.
Although these works provide some insight into how to conduct KG completion with LLMs, simply replacing PLMs with current LLMs is infeasible as finetuning LLMs on KGs is time-consuming and takes many computational resources.

With the emergence of LLMs, several works attempt to adapt LLMs for KG completion.
KG-LLM \cite{KG-LLM} performs instruction tuning on KG completion tasks with relatively smaller LLMs (e.g., LLaMA-7B, ChatGLM-6B) and surpasses ChatGPT and GPT-4, but it still lags behind state-of-the-art KG completion models.
KICGPT \cite{KICGPT} employs an embedding-based model as the retriever to generate an ordered candidate entity list and designs an in-context learning strategy to prompt ChatGPT to re-rank the entities with a multi-round interaction.
KICGPT is the most similar work to our proposed method \modelname, because we also employ an embedding-based model to obtain candidate entities and provide them to LLMs.
However, accessing closed-source LLMs like ChatGPT is costly as the inference cost grows linearly with the number of missing facts.
In contrast, we propose an effective and efficient method to finetune open-source LLMs.

\section{The \modelname Framework}\label{sect:method}
In this section, we describe the proposed \modelname framework for KG completion.

\begin{figure}[!t]
\centering  
\includegraphics[width=0.999\linewidth]{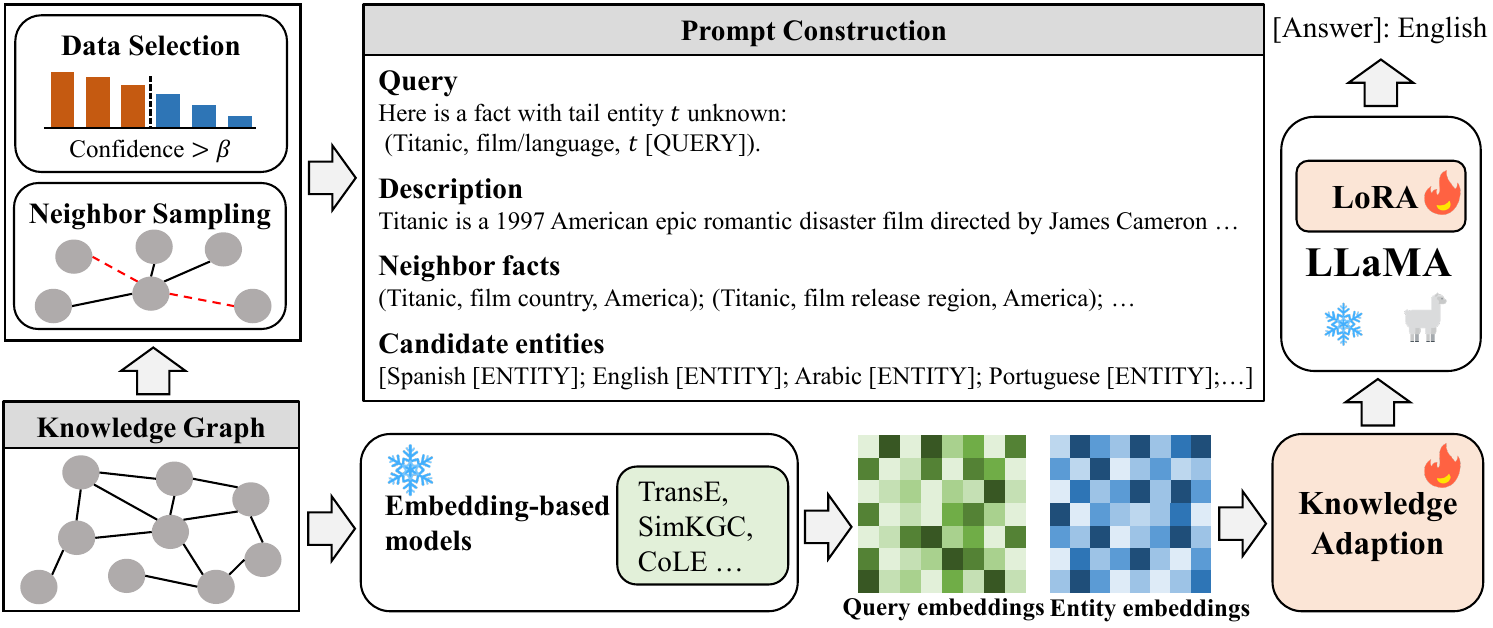}  
\caption{Illustration of the proposed \modelname framework.}  
\label{fig:framework}  
\end{figure}

\subsection{Notations}
We start by introducing the definitions and notations used in this paper.

\medskip\noindent\textbf{Knowledge graph.}
A KG is denoted as $\mathcal{G}=(\mathcal{E}, \mathcal{R}, \mathcal{T})$. 
$\mathcal{E}$ is the set of entities, and $\mathcal{R}$ is the set of relations.
$\mathcal{T}=\{(\textit{h},\textit{r}, \textit{t}) | \textit{h},\textit{t} \in \mathcal{E}, \textit{r} \in \mathcal{R}\}$ is the set of facts.
We denote a fact as $(\textit{h},\textit{r}, \textit{t})$, in which $\textit{h}$ is the head entity, $\textit{t}$ is the tail entity, and $\textit{r}$ is the relation between $\textit{h}$ and $\textit{t}$.
Furthermore, the available text attributes of $\mathcal{G}$ encompass entity names, relation names, and entity descriptions.

\medskip\noindent\textbf{Knowledge graph completion.}
KG completion (a.k.a. link prediction) aims to predict the missing entity given an incomplete fact.
To be more specific, given an incomplete fact $(\textit{h}, \textit{r}, ?)$ or $(?, \textit{r}, \textit{t})$,  the purpose of KG completion is to find the missing entity $\textit{t}$ or $\textit{h}$ from the entity set $\mathcal{E}$.

\subsection{Framework Overview}
Fig.~\ref{fig:framework} shows the overall framework of the proposed \modelname.
In general, \modelname finetunes an LLM $\mathcal{M}$ on a given KG $\mathcal{G}$ with the help of an embedding-based model $\mathcal{M}_{E}$ which has been trained on $\mathcal{G}$ in advance.
To elaborate, take a tail prediction query $q=(\textit{h},\textit{r}, ?)$ as an example, we feed $q$ into $\mathcal{M}_{E}$ to get the top-$m$ predicted entities $\mathcal{C}=[\textit{e}_1, \textit{e}_2, \dots, \textit{e}_m]$ where $m$ is a predefined hyperparameter.
Subsequently, we construct a discrimination instruction $\mathcal{P}(q)$ based on the query $q$ and the candidate entities $\mathcal{C}$.
Finally, $\mathcal{P}(q)$ is fed into $\mathcal{M}$ to select the most plausible entity.
In this way, we ensure that $\mathcal{M}$ always predicts an entity in $\mathcal{E}$ as the answer, avoiding grounding the unconstrained output texts from $\mathcal{M}$ to entities.
For efficient finetuning, we employ $\mathcal{M}_{E}$ to score the instruction samples and only keep samples with high confidence.
Additionally, to enhance the graph reasoning ability of $\mathcal{M}$, we design a knowledge adaption module to inject the embeddings of $q$ and candidate entities in $\mathcal{C}$ obtained from $\mathcal{M}_{E}$ into $\mathcal{M}$.

\subsection{Instruction Construction}
For a query $q = (\textit{h},\textit{r}, ?)$, we construct the prompt $\mathcal{P}$ by integrating four pieces of information: Query $\mathcal{Q}$, Description $\mathcal{D}$, Neighbor facts $\mathcal{N}$ and Candidate entities $\mathcal{C}$, which can be represented as:
\begin{align}
\mathcal{P}(q) = [\mathcal{Q} ; \mathcal{D} ;\mathcal{N} ; \mathcal{C}],
\end{align}
where ``$;$'' is the concatenation operation between texts.
We give an example of querying $(\textit{Titanic}, \textit{film language}, ?)$, as illustrated in Fig.~\ref{fig:framework}.

\paragraph{Query} refers to a natural language sentence containing the incomplete fact $(\textit{h},\textit{r}, ?)$.
Instead of designing a complex natural language question to prompt off-the-shelf LLMs, we simply concatenate the entity and relation names in the form of a triplet and indicate which entity is missing.
During the finetuning process, the LLM $\mathcal{M}$ will be trained to fit our prompt format.

\paragraph{Description} is the descriptive text of $h$, which contains abundant information about the entity.
This additional information helps the LLM $\mathcal{M}$ get a better understand of the entity $h$.
For instance, we depict \textit{Titanic} in Fig.~\ref{fig:framework} as \textit{a 1997 American epic romantic disaster film directed by James Cameron}.

\paragraph{Neighbor facts} are obtained by sampling facts related to the entity $h$.
As there may be numerous facts associated with $h$, we devise a straightforward yet effective sampling mechanism, namely \underline{r}elation \underline{c}o-occurrence (\emph{RC}) sampling.
It is rooted in relation co-occurrence, and streamlines the number of facts while ensuring the inclusion of relevant information.
The intuition behind RC sampling lies in the observation that the relations frequently co-occurring with $\textit{r}$ are considered to be crucial to complete $(\textit{h},\textit{r}, ?)$.
For example, the relations \textit{film language} and \textit{film\ country} in Fig.~\ref{fig:framework} often co-occur, 
because the language of a film is closely related to the country where it is released.
Therefore, we can infer that the language of \textit{Titanic} is highly likely to be English from the fact that it is an American film.
Drawing on the above observation, we sort the neighboring relations of $h$ based on their frequency of co-occurrences with $\textit{r}$ and subsequently select facts containing these relations until a preset threshold $\gamma$ is reached.

\paragraph{Candidate entities} are the names of top-$m$ entities ranked by the KG embedding model $\mathcal{M}_{E}$.
We retain the order of candidate entities since the order reflects the confidence of each entity from $\mathcal{M}_{E}$.
We instruct the LLM $\mathcal{M}$ to select an entity from the given candidates, thereby avoiding the grounding errors.

\subsection{Truncated Sampling}
\label{sec:truncated_sampling}
We design a sampling method to select representative samples to reduce instruction data.
The main idea is to opt for high-confidence samples indicated by $\mathcal{M}_{E}$, thereby empowering $\mathcal{M}$ to acquire intrinsic semantic knowledge of $\mathcal{M}_{E}$ efficiently.
By finetuning $\mathcal{M}$ on these selected instruction samples, we effectively mitigate the computational burden associated with training. 

We take the sample fact $(\textit{h},\textit{r},\textit{t})$ with query $(\textit{h},\textit{r}, ?)$ and answer entity $\textit{t}$ as an example. 
We denote the sample fact as $s$.
Specifically, we assess the confidence of $s$ from both global and local perspectives.
The global confidence $Conf_\text{global}(s)$ is computed as $\frac{1}{R(\textit{h},\textit{r}, \textit{t})}$, where $R(\textit{h},\textit{r}, \textit{t})$ is the ranking of $\textit{t}$ for the query $(\textit{h},\textit{r}, ?)$.
We name it as the global confidence because it measures the ranking of $\textit{t}$ among all candidates in the KG.

Considering that the global confidence ignores the differences between two queries whose answer entities are in the same rank, inspired by \cite{confidence}, we present the local confidence to measure the score of a fact itself.
The local confidence $Conf_\text{local}(s)$ is computed as $f(\textit{h},\textit{r}, \textit{t})$, i.e., the score of $s$ obtained from $\mathcal{M}_{E}$.
It is worth noting that $Conf_\text{local}(s)$ is assigned as 0 if $\textit{t}$ is not ranked within the top-$m$.
Finally, the confidence of $s$ is determined by the weighted sum of global and local confidence, expressed as follows:
\begin{align}
Conf(s) = Conf_\text{global}(s) + \alpha \times Conf_\text{local}(s),
\end{align}
where $\alpha$ serves as a hyperparameter to balance the global and local confidence.
Subsequently, we introduce a threshold $\beta$ and keep the samples with confidence greater than $\beta$ as the final instruction data.

\subsection{Instruction Tuning with Knowledge Adaption}
Given the prompt $\mathcal{P}(q)$, we finetune the LLM $\mathcal{M}$ to generate the entity name of $\textit{t}$.
The loss of instruction tuning is a re-construction loss:
\begin{equation}
    \mathcal{L}_{\mathcal{M}} = - \sum_{i=1}^{N} \text{log}\ p\big(y_i\,|\,y_{<i},\mathcal{P}(q)\big),
\end{equation}
where $N$ denotes the number of tokens in the entity name of $t$, $y_i\ (i=1, 2, \dots, N)$ represents the $i$-th token, and $p(y_i\,|\,y_{<i},\mathcal{P}(q))$ signifies the probability of generating $y_i$ with $\mathcal{M}$ given the prompt $\mathcal{P}(q)$ and tokens that have been generated. 

The facts provided in $\mathcal{P}(q)$ are presented in the text format, losing the global structure information of KGs.
Therefore, we propose to inject the embeddings learned from KG structure into $\mathcal{M}$ to further improve its graph reasoning ability.
We align the embeddings from $\mathcal{M}_E$ with the semantic space of $\mathcal{M}$, to get the knowledge representations:
\begin{equation}
    \hat{\mathbf{e}} = \mathbf{W}_2\big(\text{SwiGLU}(\mathbf{W}_1 \cdot \mathbf{e} + \mathbf{b}_1)\big) + \mathbf{b}_2,
\end{equation}
where $\hat{\mathbf{e}}$ denotes the knowledge representation obtained based on the embeddings $\mathbf{e}$. 
$\mathbf{W}_1 \in \mathbbm{R}^{d_1 \times d_0}$, $\mathbf{b}_1 \in \mathbbm{R}^{d_1}$, $\mathbf{W}_2 \in \mathbbm{R}^{d_2 \times d_1}$, and $\mathbf{b}_2 \in \mathbbm{R}^{d_2}$ are trainable weights. 
$d_0$ is the embedding dimension of $\mathcal{M}_E$, $d_2$ is the hidden size of $\mathcal{M}$, and $d_1$ is a hyperparameter. 
SwiGLU is a common activation function used in LLaMA \cite{LLaMA}.

Considering that $\mathcal{M}_E$ scores a fact based on the embeddings of the query $q$ and the candidate entity $\textit{t}$, we inject the knowledge representations of $q$ and all candidate entities in $\mathcal{C}$ into $\mathcal{M}$.
We add two special placeholders ``[QUERY]'' and ``[ENTITY]'' to indicate that there will be a knowledge representation from $\mathcal{M}_E$, as shown in Fig.~\ref{fig:framework}.
Specifically, we place a ``[QUERY]'' after the missing entity in $\mathcal{Q}$ and an ``[ENTITY]'' after each entity name in $\mathcal{C}$.

% ================================ Experiments =======================================================

\section{Experiments}\label{sect:experiment}

\begin{table}[!t]
\caption{The statistics of datasets.}
\centering\small
\setlength{\tabcolsep}{2pt}
\resizebox{0.8\textwidth}{!}
{
\begin{tabular}{lrrrrr}
\toprule
Datasets & \#Entities & \#Relations & \#Training & \#Validation & \#Testing \\
\midrule
FB15K-237 & 14,541 & 237 & 272,115 & 17,535 & 20,466 \\
WN18RR & 40,943 & 11 & 86,835 & 3,034 & 3,134 \\
\bottomrule
\end{tabular}}
\label{tab:data_statistics}
\end{table}

\subsection{Experiment Setup}

\noindent\textbf{Datasets.}
In the experiments, we use two benchmark datasets, \fb \cite{FB15K237} and \wn \cite{WN18RR}, to evaluate our proposed framework.
\fb consists of real-world named entities and their relations, constructed based on Freebase \cite{Freebase}.
On the other hand, \wn contains English phrases and the semantic relations between them, constructed based on WordNet \cite{WordNet}.
Notably, these two datasets are updated from their previous versions (i.e., FB15K and WN18 \cite{TransE}) respectively,
they both removed some inverse edges to prevent data leakage.
For a detailed overview, the statistics of these two datasets are shown in Table~\ref{tab:data_statistics}.

\medskip\noindent\textbf{Evaluation protocol.}
For each test fact, we conduct both head entity prediction and tail entity prediction by masking the corresponding entities, respectively.
The conventional metrics are ranking evaluation metrics, i.e., Hits@$k$ ($k=1,3,10$) and mean reciprocal rank (MRR).
Hits@$k$ is the percentage of queries whose correct entities are ranked within the top-$k$, and MRR measures the average reciprocal ranks of correct entities.
In our framework, the finetuned LLM selects an entity as the answer from the ranking list of candidates.
To assess its performance and make the results comparable to existing work,
we move the selected entity to the top of the ranking list, and other candidates remain unchanged.
We then use Hits@$k$ and MRR to assess the reranked candidate list.
We report the averaged results of head and tail entity prediction under the filtered ranking setting \cite{TransE}. 

\medskip\noindent\textbf{Implementation details.}
We run our experiments on two Intel Xeon Gold CPUs, an NVIDIA RTX A6000 GPU, and Ubuntu 18.04 LTS.
Text attributes are taken from KG-BERT \cite{KG-BERT}.
We select three representative embedding-based models to experiment with \modelname, namely, TransE, SimKGC, and CoLE.
Each embedding-based model is pre-trained on the training set. 
We obtain the top 20 predicted entities for each query in the validation set and test set.
We also obtain the embeddings of all queries and entities for knowledge adaption.

As for the instruction tuning, we select LLaMA-2-7B\footnote{\url{https://huggingface.co/meta-llama/Llama-2-7b-chat-hf}} as the LLM.
We employ LoRA \cite{LoRA} for parameter-efficient finetuning.
The hyperparameters of LoRA are set to $r=64$, $\text{alpha}=16$, and $\text{dropout}=0.1$.
We introduce LoRA for all query and value projection matrices in the self-attention module of Transformer.
To further speed up the finetuning process, we quantize the LLM by QLoRA~\cite{QLoRA}, which quantizes the LLM parameters to 4 bits by introducing Double Quantization with 4-bit NormalFloat data type.
Inspired by KICGPT~\cite{KICGPT}, we partition the validation set into two parts according to 9:1.
The first part is used to finetune the LLM to follow the instructions, and the second part is used for hyperparameter selection.
Note that we do not use the training data of each benchmark to construct instructions.
Since the embedding-based model has learned the training data, it would rank the correct entity at the first in the candidate list for most training facts.
If we use these candidate lists to construct instructions, the LLM would learn a tricky solution to pick the first candidate as an answer, which is not the goal of our finetuning.

\begin{table*}[!t]
\caption{
Link prediction Results. We mark the best scores in terms of each metric in {\bf bold} and the second-best scores are \underline{underlined}.
We reproduce the results of TransE, SimKGC and CoLE using their source code and hyperparameters.
The results of Neural-LP are obtained from \protect\cite{RNNLogic}. 
The results of GenKGC, KGT5 and KG-S2S are obtained from \protect\cite{KG-S2S}. 
The results of other baselines are taken from their respective original papers.
}
\label{tab:main_results}
\centering
\setlength{\tabcolsep}{3pt}
\small
\resizebox{\textwidth}{!}
{
\begin{tabular}{lcccccccc}
\toprule
\multirow{2}{*}{Models} & \multicolumn{4}{c}{\fb} & \multicolumn{4}{c}{\wn} \\
\cmidrule(lr){2-5} \cmidrule(lr){6-9} 
% \cline{2-5} \cline{6-9}
& MRR & Hits@1 & Hits@3 & Hits@10 & MRR & Hits@1 & Hits@3 & Hits@10 \\
\midrule
\multicolumn{9}{c}{\textbf{Embedding-based}} \\
\midrule
TransE & 0.312 & 0.212 & 0.354 & 0.510 & 0.225 & 0.016 & 0.403 & 0.521 \\
RotatE & 0.338 & 0.241 & 0.375 & 0.533 & 0.476 & 0.428 & 0.492 & 0.571 \\
TuckER & 0.358 & 0.266 & 0.394 & 0.544 & 0.470 & 0.443 & 0.482 & 0.526 \\
Neural-LP & 0.237 & 0.173 & 0.259 & 0.361 & 0.381 & 0.368 & 0.386 & 0.408 \\
NCRL & 0.300 & 0.209 & - & 0.473 & 0.670 & 0.563 & - & \bf{0.850} \\
CompGCN & 0.355 & 0.264 & 0.390 & 0.535 & 0.479 & 0.443 & 0.494 & 0.546 \\
HittER & 0.373 & 0.279 & 0.409 & 0.558 & 0.503 & 0.462 & 0.516 & 0.584 \\
NBFNet & \underline{0.415} & 0.321 & \underline{0.454} & \bf{0.599} & 0.551 & 0.497 & 0.573 & 0.666 \\
\midrule
KG-BERT & - & - & - & 0.420 & 0.216 & 0.041 & 0.302 & 0.524 \\
StAR & 0.365 & 0.266 & 0.404 & 0.562 & 0.551 & 0.459 & 0.594 & 0.732 \\
MEM-KGC & 0.346 & 0.253 & 0.381 & 0.531 & 0.557 & 0.475 & 0.604 & 0.704 \\
SimKGC & 0.338 & 0.252 & 0.364 & 0.511 & \underline{0.671} & \underline{0.595} & \underline{0.719} & 0.802 \\
CoLE & 0.389 & 0.294 & 0.429 & 0.572 & 0.593 & 0.538 & 0.616 & 0.701 \\
\midrule
\multicolumn{9}{c}{\textbf{Generation-based}} \\
\midrule
GenKGC & - & 0.192 & 0.355 & 0.439 & - & 0.287 & 0.403 & 0.535 \\
KGT5 & 0.276 & 0.210 & - & 0.414 & 0.508 & 0.487 & - & 0.544 \\
KG-S2S & 0.336 & 0.257 & 0.373 & 0.498 & 0.574 & 0.531 & 0.595 & 0.661 \\
$\text{ChatGPT}_\text{one-shot}$ & - & 0.267 & - & - & - & 0.212 & - & - \\
KICGPT & 0.412 & 0.327 & 0.448  & 0.581 & 0.564 & 0.478 & 0.612 & 0.677 \\
\midrule
{LLaMA + TransE} & 0.232 & 0.080 & 0.321 & 0.502 & 0.202 & 0.037 & 0.360 & 0.516 \\
{LLaMA + SimKGC} & 0.236 & 0.074 & 0.335 & 0.503 & 0.391 & 0.065 & 0.695 & 0.798 \\
{LLaMA + CoLE} & 0.238 & 0.033 & 0.387 & 0.561 & 0.374 & 0.117 & 0.602 & 0.697 \\
{\modelname + TransE} & 0.389 & 0.322 & 0.408 & 0.525 & 0.491 & 0.462 & 0.496 & 0.560 \\
{\modelname + SimKGC} & 0.402 & \underline{0.338} & 0.418 & 0.528 & \bf{0.686} & \bf{0.616} & \bf{0.730} & \underline{0.806} \\
{\modelname + CoLE} & \bf{0.439} & \bf{0.364} & \bf{0.468} & \underline{0.586} & 0.617 & 0.569 & 0.638 & 0.708 \\
\bottomrule
\end{tabular}}
\end{table*}

\begin{table*}[!t]
\caption{Results of ablation study}
\label{tab:ablation_study}
\centering
\setlength{\tabcolsep}{3pt}
\resizebox{\textwidth}{!}
{
\begin{tabular}{lcccccccc}
\toprule
 \multirow{2}{*}{Models} & \multicolumn{4}{c}{\fb} & \multicolumn{4}{c}{\wn} \\
%\cline{2-5} \cline{6-9}
\cmidrule(lr){2-5} \cmidrule(lr){6-9}
& MRR & Hits@1 & Hits@3 & Hits@10 & MRR & Hits@1 & Hits@3 & Hits@10 \\
\midrule
\modelname & \bf{0.439} & \bf{0.364} & 0.468 & 0.586 & \bf{0.617} & \bf{0.569} & \bf{0.638} & 0.708 \\
\ \ w/o truncated sampling & 0.423 & 0.338 & 0.459 & 0.587 & 0.600 & 0.537 & \bf{0.638} & \bf{0.712} \\
\ \ w/o RC sampling & 0.434 & 0.354 & 0.468 & \bf{0.588} & 0.614 & 0.564 & 0.636 & 0.708 \\
\ \ w/o description & 0.436 & 0.358 & 0.467 & 0.586 & 0.603 & 0.548 & 0.630 & 0.705 \\
\ \ w/o neighbors & 0.438 & 0.360 & \bf{0.469} & \bf{0.588} & 0.610 & 0.558 & 0.637 & 0.708 \\
\ \ w/o knowledge adaption & 0.437 & 0.358 & 0.468 & 0.587 & 0.612 & 0.560 & 0.637 & 0.708 \\
\bottomrule
\end{tabular}}
\end{table*}

\subsection{Baselines}

\noindent\textbf{Embedding-based models.}
We choose eight \textit{structure-based models} as baselines.
Three triplet-based models are selected, including TransE \cite{TransE}, RotatE \cite{RotatE}, and TuckER \cite{TuckER}.
We also choose two path-based models.
Neural-LP \cite{Neural-LP} is the first model that learns logic rules from relation paths and NCLR \cite{NCRL} is the state-of-the-art path-based model. 
The remaining models are all graph-based.
CompGCN \cite{CompGCN} employs GCNs to encode the multi-relational graph structure of the KG, 
while HittER \cite{HittER} leverages the Transformer architecture.
NBFNet \cite{NBFNet} currently performs best among the structure-based models.
We also select five \textit{PLM-based models} as the competitors, namely KG-BERT \cite{KG-BERT}, StAR \cite{StAR}, MEM-KGC \cite{MEM-KGC}, SimKGC \cite{SimKGC}, and CoLE \cite{CoLE}.
Note that,
SimKGC stands the state-of-the-art link prediction model on \wn, which benefits from efficient contrastive learning.
CoLE promotes PLMs and structure-based models mutually to achieve the best performance on \fb among PLM-based models.
To ensure a fair comparison, we present results derived solely from N-BERT, the PLM-based KG embedding module within CoLE, rather than the entire CoLE framework.

\medskip\noindent\textbf{Generation-based models.}
We select three generation-based KG completion models, all of which are either based on BART or T5, namely, GenKGC~\cite{GenKGC}, KGT5~\cite{KGT5}, and KG-S2S~\cite{KG-S2S}.
Further, we select two recent models based on LLMs as baselines.
$\text{ChatGPT}_\text{one-shot}$ is a baseline proposed by AutoKG \cite{AutoKG}, and KICGPT evaluates it on the whole test sets of \fb and \wn for comparison.
KICGPT is the most competitive KG completion model, which employs RotatE to provide the top-$m$ predicted entities for each query and re-ranks these candidates with ChatGPT through multi-round interactions.
We also report the performance of \modelname without finetuning, denoted by LLaMA+TransE, LLaMA+SimKGC, and LLaMA+CoLE, respectively.

\subsection{Main Results}
\label{sec:main_results}

We report the link prediction results on \fb and \wn in Table~\ref{tab:main_results}.
Generally speaking, our proposed framework \modelname achieves the best performance in most metrics on two datasets.
Compared with the selected embedding-based models TransE, SimKGC, and CoLE, \modelname improves the performance of these models on both datasets, significantly in terms of Hits@1.
Without finetuning, the performance of \modelname drops dramatically, which demonstrates that it is necessary to finetune the LLM for KG completion task.
 
Compared with the LLM-based model $\text{ChatGPT}_\text{one-shot}$, \modelname consistently outperforms it in terms of Hits@1, regardless of the integration with any of the embedding-based models.
This indicates that prompting ChatGPT with in-context learning is less effective than finetuning a smaller LLM with the help of existing embedding-based models for link prediction.
Compared with the most competitive baseline model KICGPT which also provides the LLM with candidate entities, 
the relative improvement brought by \modelname is less. 
However, KICGPT needs multi-round interactions with ChatGPT, which has 175B parameters.
In contrast, \modelname finetunes a small LLaMA with only 7B parameters.

Comparing different metrics, we find that the performance improvement is more significant on Hits@1 while less significant on Hits@10.
In \modelname, we ask the LLM to select the plausible entity from the given candidate list.
Given that the correct entity is more likely to be ranked in the top 10 entities rather than outside the top 10, the LLM is more likely to select an entity in the top 10 as the answer.
Thus, the improvement is more obvious on Hits@1 rather than Hits@10.

We also find that the performance improvement on \fb is more significant than that on \wn. 
This discrepancy can be attributed to the stark disparity in density between the two datasets: \fb is considerably denser than \wn, implying a richer reservoir of knowledge.
More knowledge leads to better improvement since the knowledge is provided for the LLM to comprehend in the form of prompts and embeddings.

\subsection{Ablation Study}
For the ablation study, we select CoLE as the embedding-based model to provide candidate entities since \modelname with CoLE performs best overall on both datasets.
We evaluate the effectiveness of two kinds of sampling mechanisms, i.e., \emph{truncated sampling} and \emph{RC sampling}, as well as three kinds of support information, i.e., \emph{description}, \emph{neighbors}, and embeddings used in \emph{knowledge adaption}.

From the results presented in Table~\ref{tab:ablation_study}, it is evident that all components contribute a lot to \modelname.
Among all these components, \datasampling has the most substantial impact on performance.
The Hits@1 score experiences a degradation of at least 5.6\% in the absence of \datasampling.  
This shows that this mechanism can effectively select useful instruction data for the LLM to learn intrinsic semantic knowledge of the embedding-based model.

We can also observe that the impact of description, neighbors, and RC sampling differs significantly between the two datasets.
Without description, the Hits@1 will drop more on \wn.   
This is attributed to \wn being a sparse KG with less structural information compared with \fb.
Therefore, it needs additional description to enrich entity information, aiding in the differentiation between similar entities.
In addition, neighbor information is also more important for \wn.
This is because many correct entities will directly appear in the neighbor facts of \wn, facilitating the LLM in making accurate predictions.
Instead, the improvement of Hits@1 is more significant on \fb than \wn for RC sampling.
We posit that this is attributed to \fb being highly dense, with each entity having numerous neighbor facts.
Many of these facts are irrelevant to the query, leading to interference. 
Hence, RC sampling can minimize irrelevant facts and enhance effectiveness.

As for knowledge adaption, we observe consistent performance improvements across the two datasets, indicating good generality and robustness.

\begin{figure}[t]  
\centering  
\includegraphics[width=0.95\columnwidth]{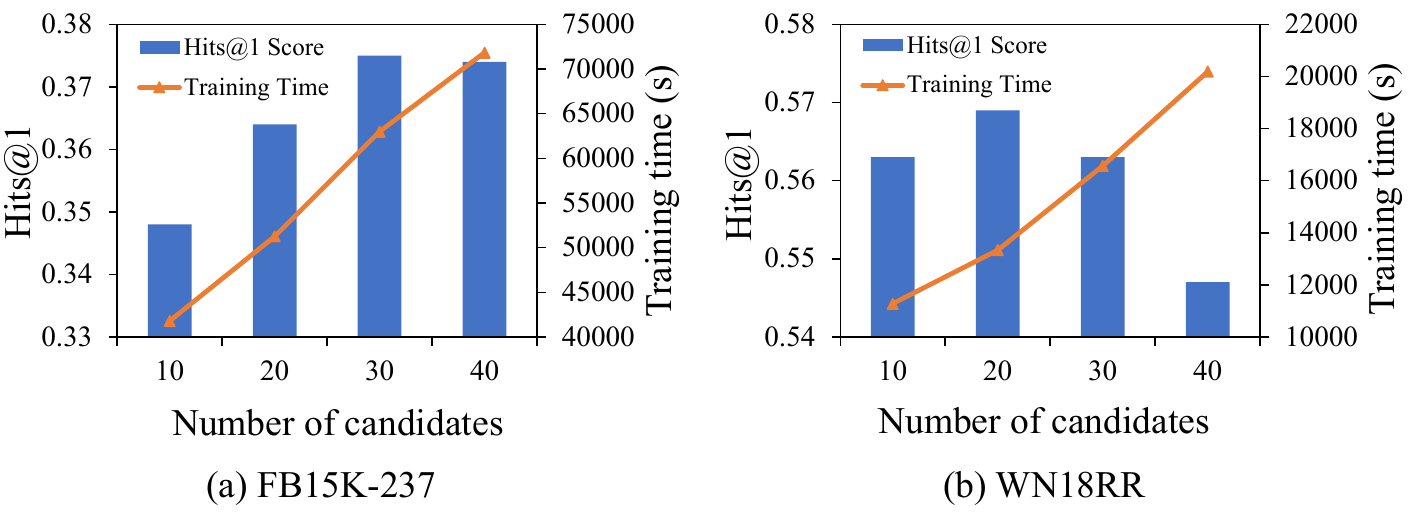}  
\caption{Hits@1 results and training time of \modelname on \fb and \wn along with the numbers of candidate entities.}  
\label{fig:candidate_number}  
\end{figure}

\subsection{Further Analyses}

\noindent\textbf{Effect of the number of candidates.}
In Section~\ref{sec:main_results}, we set the number of candidate entities $m$ provided by the embedding-based model to 20.
Here we investigate the effect of $m$ on the performance and the training time of \modelname.
The results are shown in Fig.~\ref{fig:candidate_number}.
First, for the training time, we find that it grows linearly when we increase $m$.
It is intuitive since increasing $m$ leads to longer prompts.
Second, as for the performance of \modelname, we find that the performance is best when $m$ is set to 30 on \fb, and there is a slight drop when $m$ is set to 40.
The same observation can be found on \wn if we continue to increase $m$ after 20.
This indicates that blindly increasing the number of candidate entities cannot improve performance.
Third, we find that the performance is best when $m$ is set to 30 on \fb and 20 on \wn.
That is to say, to achieve the best performance, \modelname needs more candidate entities on \fb than \wn.
We think that this discrepancy arises from the generally inferior performance of models on \fb compared to \wn.
Consequently, to ensure the presence of answer entities within the prompts, a larger $m$ is advisable on \fb than on \wn.

\begin{figure}[t]
\centering  
\includegraphics[width=0.95\columnwidth]{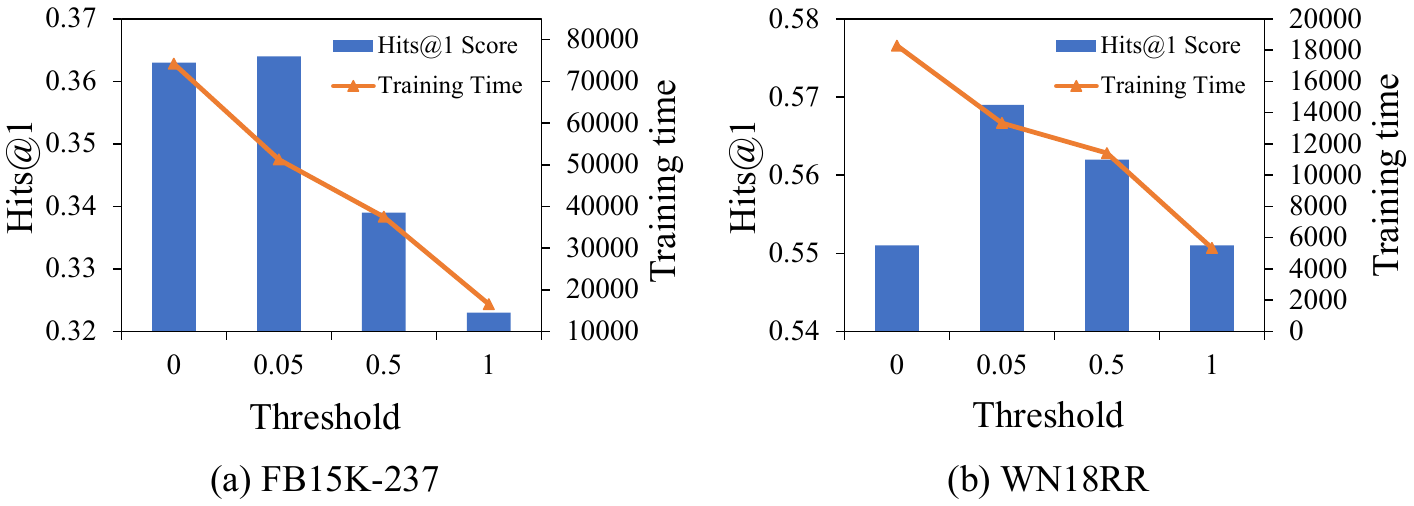}  
\caption{Hits@1 results and training time of \modelname on \fb and \wn along with the threshold for truncated sampling.}  
\label{fig:sampling_threshold}  
\end{figure}

\medskip\noindent\textbf{Effect of truncated sampling thresholds.}
In Section~\ref{sec:truncated_sampling}, we use a threshold $\beta$ to control the quantity of instruction data.
To investigate the impact of $\beta$ on the performance and the training time of \modelname, we conduct an experiment by setting different values for $\beta$.
In particular, we change $\beta$ from 0.05 in the main experiments to 0, 0.5, and 1.0 respectively.
The results are shown in Fig.~\ref{fig:sampling_threshold}. We have the following observations.
First, with $\beta$ increasing, the quantity of instruction data decreases, and therefore the training time also decreases accordingly.
Second, the performance drops when we set $\beta$ to 0 on both datasets, which indicates that increasing the quantity of instruction data does not necessarily improve the performance, and its quality also affects the performance.
Third, if we strictly ensure that the quality of the instruction data is high enough, i.e., we set $\beta$ to 0.5 or 1.0, the performance of \modelname also drops.
We think there are mainly two reasons:
(1) When $\beta$ is set to 0.5 or 1.0, the limited instruction data is not enough to finetune the LLM sufficiently.
(2) Instruction data with high confidence usually places the answer entity among the first few in the candidate list.
Therefore, finetuning the LLM with this data will cause the LLM to always choose the top-ranked entities, regardless of whether they are correct.

\begin{figure}[t]
\centering  
\includegraphics[width=.65\columnwidth]{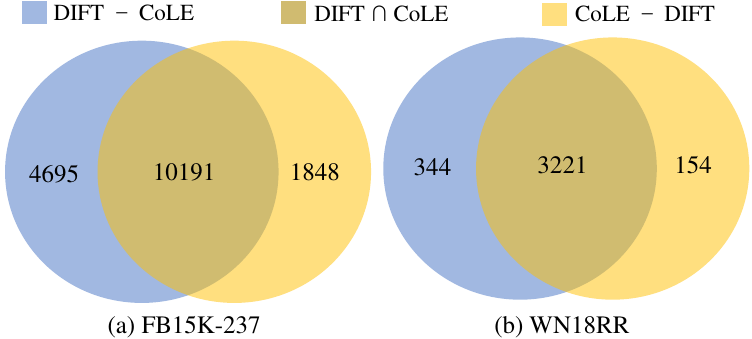}  
\caption{
Correct predictions of \modelname and CoLE on \fb and \wn.
The light blue area represents the accurate triplets predicted by \modelname, excluding those that can also be predicted by CoLE.
The dark green area illustrates the overlapping triplets predicted accurately by \modelname and CoLE.
The light green area represents the accurate triplets predicted by CoLE, excluding those that can also be predicted by \modelname.
}  
\label{fig:res_compare}  
\end{figure}

\medskip\noindent\textbf{Comparison of \modelname and basic embedding models.}
We further investigate the predictions of \modelname in comparison with those of the selected embedding-based model.
For this analysis, we continue to employ CoLE as the embedding-based model to analyze the results.
We draw Venn diagrams to highlight both their shared and individual correct predictions, as illustrated in Fig.~\ref{fig:res_compare}.
It is obvious that in addition to the shared correct predictions, \modelname can also get some correct predictions by itself. 
Conversely, we observe instances where CoLE makes correct inferences that \modelname fails to replicate.
Based on the divergence between the correct predictions of \modelname and CoLE, we can conclude that the LLM does not repeat the predicted entities by CoLE blindly, instead, it can reason the missing facts based on its knowledge obtained in the pre-training stage.

\begin{table*}[!t]
\caption{
Link prediction Results of different versions of LLaMA-2-7B.
}
\label{tab:llm_version}
\centering
\setlength{\tabcolsep}{3pt}
\small
\resizebox{\textwidth}{!}
{
\begin{tabular}{lcccccccc}
\toprule
\multirow{2}{*}{Models} & \multicolumn{4}{c}{\fb} & \multicolumn{4}{c}{\wn} \\
\cmidrule(lr){2-5} \cmidrule(lr){6-9} 
& MRR & Hits@1 & Hits@3 & Hits@10 & MRR & Hits@1 & Hits@3 & Hits@10 \\
\midrule
\multicolumn{9}{c}{\textbf{LLaMA-2-7B-Chat}} \\
\midrule
{\modelname + TransE} & 0.389 & 0.322 & 0.408 & 0.525 & 0.491 & 0.462 & 0.496 & 0.560 \\
{\modelname + SimKGC} & 0.402 & 0.338 & 0.418 & 0.528 & 0.686 & 0.616 & 0.730 & 0.806 \\
{\modelname + CoLE} & 0.439 & 0.364 & 0.468 & 0.586 & 0.617 & 0.569 & 0.638 & 0.708 \\
\midrule
\multicolumn{9}{c}{\textbf{LLaMA-2-7B-Foundation}} \\
\midrule
{\modelname + TransE} & 0.393 & 0.328 & 0.409 & 0.525 & 0.481 & 0.450 & 0.486 & 0.552 \\
{\modelname + SimKGC} & 0.405 & 0.341 & 0.420 & 0.530 & 0.682 & 0.608 & 0.731 & 0.806 \\
{\modelname + CoLE} & 0.439 & 0.363 & 0.468 & 0.587 & 0.619 & 0.571 & 0.641 & 0.710 \\
\bottomrule
\end{tabular}}
\end{table*}

\medskip\noindent\textbf{Comparison of different versions of the LLM.}
In the main experiment, we employ LLaMA-2-7B-Chat as the LLM for \modelname.
In order to investigate the influence of different versions of the LLM on the performance of \modelname, we experiment with the foundation version, denoted as LLaMA-2-7B-Foundation.
The results are shown in Table~\ref{tab:llm_version}.
\modelname with LLaMA-2-7B-Foundation performs slightly better than that with LLaMA-2-7B-Chat on \fb, but the observation is the opposite on \wn.
Generally speaking, \modelname achieves a similar performance no matter which version of the LLM are employed.
It demonstrates the robustness and generalization of \modelname for different LLM versions.

\subsection{The Finetuning Learns What?}
In this section, we investigate what the LLM learns during our finetuning process.
\modelname employs a lightweight embedding-based model to provide candidate entities for both finetuning and inference.
A natural question arises: Does the LLM learn the preference of the embedding-based model predictions or the knowledge in the KG?
To answer this question, we design the following experiment to evaluate the effect of the candidate order in both the finetuning and inference stages.

\begin{table*}[!t]
\renewcommand{\arraystretch}{1.1}
\caption{Influence of the order of candidates}
\label{tab:candidate_order}
\centering
\setlength{\tabcolsep}{4pt}
\resizebox{\textwidth}{!}
{
\begin{tabular}{cccccccccc}
% \hline
\toprule
\multirow{2}{*}{\makecell{Ordered\\finetuning}} & \multirow{2}{*}{\makecell{Ordered\\inference}} & \multicolumn{4}{c}{\fb} & \multicolumn{4}{c}{\wn} \\
\cmidrule(lr){3-6} \cmidrule(lr){7-10} & & MRR & Hits@1 & Hits@3 & Hits@10 & MRR & Hits@1 & Hits@3 & Hits@10 \\
\midrule
\Checkmark & \Checkmark & 0.439 & 0.364 & 0.468 & 0.586 & 0.686 & 0.616 & 0.730 & 0.806 \\
\Checkmark & \XSolidBrush & 0.328 & 0.168 & 0.441 & 0.584 & 0.484 & 0.233 & 0.712 & 0.806 \\
\XSolidBrush & \Checkmark & 0.423 & 0.333 & 0.466 & 0.589 & 0.627 & 0.500 & 0.731 & 0.809 \\
\XSolidBrush & \XSolidBrush & 0.417 & 0.324 & 0.464 & 0.591 & 0.625 & 0.493 & 0.736 & 0.808 \\
\bottomrule
\end{tabular}}
\end{table*}

\medskip\noindent\textbf{Effect of the order of candidates.}
\modelname takes the top-$m$ predicted entities from the embedding-based model as the candidates for the LLM.
We retain the order of candidates because we assume that the order reflects the knowledge learned by the embedding-based model.
Here, to investigate the influence of the order of candidates, we shuffle the candidates in the finetuning or inference stages to ask the LLM to select an entity from the shuffled candidate list.
Remember that the shuffled candidate list is only used for entity selection, we move the selected entity to the top of the ranking list from the embedding-based model for evaluation.
Results are shown in Table~\ref{tab:candidate_order}, and we have the following observations.

On \fb, we employ CoLE as the embedding-based model.
We can find that the performance drops dramatically if we finetune the LLM with ordered candidates but shuffle the candidates during inference.
We think the reason is that ordered candidates instruct the LLM to select within the top few entities as they are more plausible.
Therefore, the LLM still focus on the top few candidates during inference, even though the candidates are shuffled.
When we finetune the LLM with shuffled candidates, we find that the performance changes slightly whether the candidates are shuffled or not during inference.
The reason is that the LLM has no idea about the preference that the top few candidates are more plausible, so it can not benefit from the order of candidates.

On \wn, we use SimKGC as the embedding-based model and similar observations can be found.
However, we find that the performance of \modelname is even worse than SimKGC when we finetune the LLM with shuffled candidates.
This demonstrates that the LLM can not outperform SimKGC solely based on its inherent knowledge without prediction preferences.

Based on the above analyses, it appears that our \modelname not only captures prediction preferences but also primarily acquires knowledge from the KG.

\medskip\noindent\textbf{Case study.}
To explore how \modelname improves performance compared with the selected embedding-based models, we conduct a case study on \modelname (integrating CoLE), TransE, SimKGC, and CoLE.
Table~\ref{tab:case_study} presents the Hits@1 results of the four models on three queries from \fb, in which the entities marked with horizontal lines at the bottom are the answers.
In the first two cases, \modelname consistently performs accurately while the other models all predict wrong entities.
\begin{itemize}
\item In Case 1, the contextual description of the head entity, ``\textit{It tells the story of an aspiring actress named Betty Elms, newly arrived in Los Angeles ...}'', offers ample support to determine the answer ``Los Angeles'', and our \modelname generates the correct entity name, indicating that \modelname has improved contextual inference capability compared with the embedding-based models.

\item In Case 2, neither the description nor the neighbor information provides clues to \textit{Shonda Rhimes}' gender.
It is difficult for embedding-based models to infer the correct entity based on such incomplete knowledge.
Instead, \modelname has open knowledge and powerful commonsense reasoning ability, allowing it to overcome this limitation and predict the correct answers.
This case shows the complementarity of embedding-based models and LLMs in our framework.

\item In Case 3, despite \modelname inferring an ``incorrect'' entity ``\textit{The Last King of Scotland}'', it is crucial to highlight that the underlying issue is associated with the dataset, not \modelname itself.
This is because the language of ``\textit{The Last King of Scotland}'' is also English, but \fb lacks this specific knowledge.
This case demonstrates that \modelname is capable of leveraging open knowledge in LLMs, surpassing the constraints of closed knowledge in KGs.
\end{itemize}

\begin{table}[!t]
\centering
\renewcommand{\arraystretch}{1.1}
\caption{Case study on three queries from \fb. Correct answers are \underline{underlined}. }
\label{tab:case_study}
\resizebox{1.0\textwidth}{!}
{
\setlength{\tabcolsep}{6pt}\small
    \begin{tabular}{llll} 
    \toprule
    & Case 1 & Case 2 & Case 3 \\
    \midrule
    {Head entity}
    & {Mulholland Drive} & {Shonda Rhimes} & {{?}}  \\
    \textit{Relation} & \textit{featured film locations} & \textit{gender} & \textit{film language}  \\
    {Tail entity} & {{?}} & {{?}} & {English language}  \\  
    \midrule
    \modelname & \underline{{Los Angeles}} & \underline{{Female}} &  {The Last King of Scotland}  \\  
    TransE & {Paris} & {Male} &  {Pan's Labyrinth}  \\  
    SimKGC & {Berkeley} & {Male} &  \underline{{The Illusionist}}  \\  
    CoLE & {New York City} & {Male} &  \underline{{The Illusionist}}  \\  
    \bottomrule
    \end{tabular}
}
\end{table}

\section{Conclusion and Future Work}\label{sect:conclusion}
In this paper, we propose a novel KG completion framework \modelname.
It finetunes generative LLMs with discrimination instructions using LoRA, which does not involve grounding the output of LLMs to entities in KGs.
To further reduce the computation cost and make \modelname more efficient, we propose a truncated sampling method to select facts with high confidence for finetuning.
KG embeddings are also added into the LLMs to improve the finetuning effectiveness.
Experiments show that \modelname achieves state-of-the-art results on KG completion.
In future work, we plan to support other KG tasks such as KGQA and entity alignment.

\subsubsection*{Acknowledgments}
This work was funded by National Natural Science Foundation of China (No. 62272219), Postdoctoral Fellowship Program of CPSF (No. GZC20240685), and CCF-Tencent Rhino-Bird Open Research Fund.

\subsubsection*{Supplemental Material Statement:} Source code and datasets are available on GitHub: \url{https://github.com/nju-websoft/DIFT}.

\bibliographystyle{splncs04}
\bibliography{iswc24}

\end{document}